\pdfoutput=1
\documentclass[runningheads]{llncs}
\usepackage{graphicx}
\usepackage[hidelinks]{hyperref}
\usepackage{times}
\usepackage{latexsym}

\usepackage[square,numbers]{natbib} 

\usepackage[T1]{fontenc}
\usepackage[utf8]{inputenc}
\usepackage{microtype}
\usepackage{tabularx}
\usepackage{booktabs}
\usepackage{multirow}
\usepackage{graphicx}
\usepackage{amsmath}
\usepackage{xcolor}
\usepackage{enumitem}
\usepackage[normalem]{ulem}
\newcommand{\pz}{\phantom{0}}

\begin{document}
\title{Balancing the Style-Content Trade-Off in Sentiment Transfer Using Polarity-Aware Denoising}
\titlerunning{Sentiment Transfer Using Polarity-Aware Denoising}
\author{Sourabrata Mukherjee\orcidID{0000-0002-1713-2769} 
\and Zdeněk Kasner\orcidID{0000-0002-5753-5538} 
\and Ondřej Dušek\orcidID{0000-0002-1415-1702}} 

\authorrunning{Mukherjee et al.}
\institute{Charles University, Faculty of Mathematics and Physics,\\ Institute of Formal and Applied Linguistics \\Prague, Czechia\\ \email{\{mukherjee,kasner,odusek\}@ufal.mff.cuni.cz}}

\maketitle              
\begin{abstract}
Text sentiment transfer aims to flip the sentiment polarity of a sentence  (positive to negative or vice versa) while preserving its sentiment-independent content.  Although current models show good results at changing the sentiment, content preservation in transferred sentences is insufficient. In this paper, we present a sentiment transfer model based on polarity-aware denoising, which accurately controls the sentiment attributes in generated text, preserving the content to a great extent and helping to balance the style-content trade-off. Our proposed model is structured around two key stages in the sentiment transfer process: better representation learning using a shared encoder 
and sentiment-controlled generation using separate sentiment-specific decoders. 
Empirical results show that our methods outperforms state-of-the-art baselines in terms of content preservation while staying competitive in terms of style transfer accuracy and fluency. Source code, data, and all other related details are available on Github.\footnote{\url{https://github.com/SOURO/polarity-denoising-sentiment-transfer}}


\end{abstract}
\section{Introduction}
Text sentiment transfer is the task of changing the sentiment polarity of a text while retaining sentiment-independent semantic content (e.g., ``The food was tasteless'' to ``The food was delicious'') \cite{shen2017style, prabhumoye2018style, li2018delete, luo2019towards}. It has been introduced in the context of textual style transfer, where positive and negative sentiment are considered distinct styles \cite{luo2019towards}.
Style transfer is motivated by various writing assist tasks for copywriting or personalized chatbots, e.g.\ changing review sentiment, debiasing or simplifying a news text, or removing offensive language  \cite{luo2019towards, santos2018fighting, jin_deep_2021}.

With the success of deep learning in the last decade, a variety of neural methods have been proposed for this task \cite{toshevska2021review, jin_deep_2021}. If parallel data are provided, standard sequence-to-sequence models can be directly applied \cite{rao2018dear}. However, due to lack of parallel corpora (paired sentences with opposite sentiment and otherwise identical content), learning sentiment transfer -- and text style transfer in general -- from unpaired data represents a substantial research challenge.

A first approach to learning text style transfer from unpaired data disentangles text representation  in a latent space into style-independent content and stylistic attributes (such as sentiment polarity) and applies generative modeling \cite{hu2017toward, shen2017style, prabhumoye2018style}.
The latent representation needs to preserve the meaning of the text while abstracting away from its stylistic properties, which is not trivial \cite{lample2018multiple}. In fact, disentanglement is impossible in theory without inductive biases or other forms of supervision \cite{locatello2019challenging}.
A second line of research is prototype editing \cite{li2018delete,fu_rethinking_2019}, which focuses specifically on style marker words (also called \emph{pivot} words, e.g. sentiment polarity indicating words such as “good” or “bad”). The approach typically extracts a sentence “template” without the pivots and then fills in pivots marking the target style. However, since the pivot words are typically extracted using simple unsupervised probabilistic methods, they are difficult to distinguish from content words, which again leads to content preservation errors.

Our work combines both research branches and extends them, using additional supervision. The supervision comes from a sentiment dictionary, which is applied on pivot words within the context of generative models to learn better latent representations via the task of  polarity-aware denoising.
First, we randomly delete (or mask) pivot word(s) of input sentences. Then a shared encoder pre-trained on general domain helps in preparing a latent representation, followed by separate sentiment-specific decoders that are used to change the sentiment of the original sentence. We follow back-translation for style transfer approach proposed by \citet{prabhumoye2018style} to represent the sentence meaning in the latent space. 

Our contributions are summarized as follows:
\begin{itemize}[itemsep=0pt,topsep=1pt,leftmargin=12pt]
    \item We design a sentiment transfer model using an extended transformer architecture and polarity-aware denoising. Our use of separate sentiment-specific decoders and polarity-aware denoising training allow more control over both the target sentiment and the sentiment-independent content.  
    \item We derive a new non-parallel sentiment transfer dataset from the Amazon Review Dataset \cite{ni2019justifying}. It is more topically diverse than earlier used datasets Yelp \cite{li2018delete} and IMDb \cite{dai2019style}, which were majorly focused on movie and restaurant/business-related reviews. Our dataset and code is publicly available.\footnotemark[\value{footnote}]
    \item We introduce polarity-masked BLEU (Mask\-BLEU) and similarity score (MaskSim) for automatic evaluation of content preservation in this task. These metrics are derived from the traditional BLEU score \cite{papineni2002bleu} and Sentence BERT-based cosine similarity score \cite{reimers2019sentence}. In our approach, we mask polarity words beforehand for sentiment-independent content evaluation.  
    \item Both automatic and human evaluations on our dataset show that our proposed approach generally outperforms state-of-the-art (SotA) baselines. Specifically, with respect to content preservation, our approach achieves substantially better performance than other methods. 
    \end{itemize}

\section{Related Work}

\paragraph{Sentiment Transfer} A common approach to the sentiment transfer task is to separate content and style in a latent space, and then adjust the separated style. \citet{hu2017toward} use a variational autoencoder and factor its latent representation into a style-independent and stylistic parts. \citet{fu2017style} compare a multi-decoder model with a setup using a single decoder and style embeddings. \citet{shen2017style} propose a cross-aligned auto-encoder with adversarial training. \citet{prabhumoye2018style} propose to perform text style transfer through back-translation. In a recent work, \citet{he2020probabilistic} apply variational inference. Although these approaches successfully change the text style, they also change the text content, which is a major problem. Many previous methods \cite{hu2017toward, shen2017style, fu2017style, prabhumoye2018style} formulate the style transfer using the encoder-decoder framework. The encoder maps the text into a style-independent latent representation, and the decoder generates the target text using the latent representation and a style marker. Again, a major issue of these models is poor preservation of non-stylistic semantic content.

\paragraph{Content Preservation} To further deal with the above problem, \citet{li2018delete} first extract content words by deleting phrases, then retrieve new phrases associated with the target attribute, and finally use a neural model to combine these into a final output. 
\citet{luo2019towards} employ a dual reinforcement learning framework with two sequence-to-sequence models in two directions, using style classifier and back-transfer reconstruction probability as rewards. Though these works show some improvement, they are still not able to properly balance preserving the content with transferring the style.
Our polarity-aware denoising technique aims to solve this problem by specifically targeting and changing polarity words while preserving the rest of the content (see Section \ref{sec:pnoising}).

\paragraph{Evaluation} Another challenge remains in the evaluation of textual style transfer. 
Previous work on style transfer \cite{hu2017toward, prabhumoye2018style, dai2019style, he2020probabilistic} has re-purposed evaluation metrics from other fields, such as BLEU from machine translation \cite{papineni2002bleu} and PINC from paraphrasing \cite{chen2011collecting}. However, these metric cannot evaluate style transfer specifically with respect to preservation of content \cite{toshevska2021review} as they do not take into account the necessity of changing individual words when altering the style. Intended differences between the source sentence and the transferred sentence are thus penalized. 
In this regard, we have introduced polarity masked BLEU score (MaskBLEU) and polarity masked similarity measure (MaskSim), where we have masked the polarity words beforehand (see Section \ref{sec:autom-eval}). 

\section{Approach}
\label{sec:approach}

\subsection{Task Definition}
\label{sec:task-definition}
Given two datasets, $X_{\mathit{pos}} = \{ x_{1}^{(\mathit{pos})}, \dots, x_{m}^{(\mathit{pos})} \}$ and $X_{\mathit{neg}} = \{ x_{1}^{(\mathit{neg})},\dots, x_{n}^{(\mathit{neg})}\}$ which represent two different sentiments $\mathit{pos}$ and $\mathit{neg}$, respectively, our task is to generate sentences of the desired sentiment while preserving the meaning of the input sentence. Specifically, we alter samples of dataset $X_{\mathit{pos}}$ such that they belong to sentiment $\mathit{neg}$ and samples of $X_{\mathit{neg}}$ such that they belong to sentiment $\mathit{pos}$, while sentiment-independent content is preserved. We denote the output of dataset $X_{\mathit{pos}}$ transferred to sentiment $\mathit{neg}$ as $X_{\mathit{pos} \rightarrow \mathit{neg}} = \{ \hat{x}_{1}^{(\mathit{neg})}, \dots, \hat{x}_{m}^{(\mathit{neg}}) \}$ and the output of dataset $X_{\mathit{neg}}$ transferred to sentiment $\mathit{pos}$ as $X_{\mathit{neg} \rightarrow \mathit{pos}} = \{ \hat{x}_{1}^{(\mathit{pos})},\dots, \hat{x}_{n}^{(\mathit{pos})}\}$.

\subsection{Model Overview}
\label{sec:model-overview}

\begin{figure*}[t]
    \centering
    \includegraphics[width=1.0\textwidth]{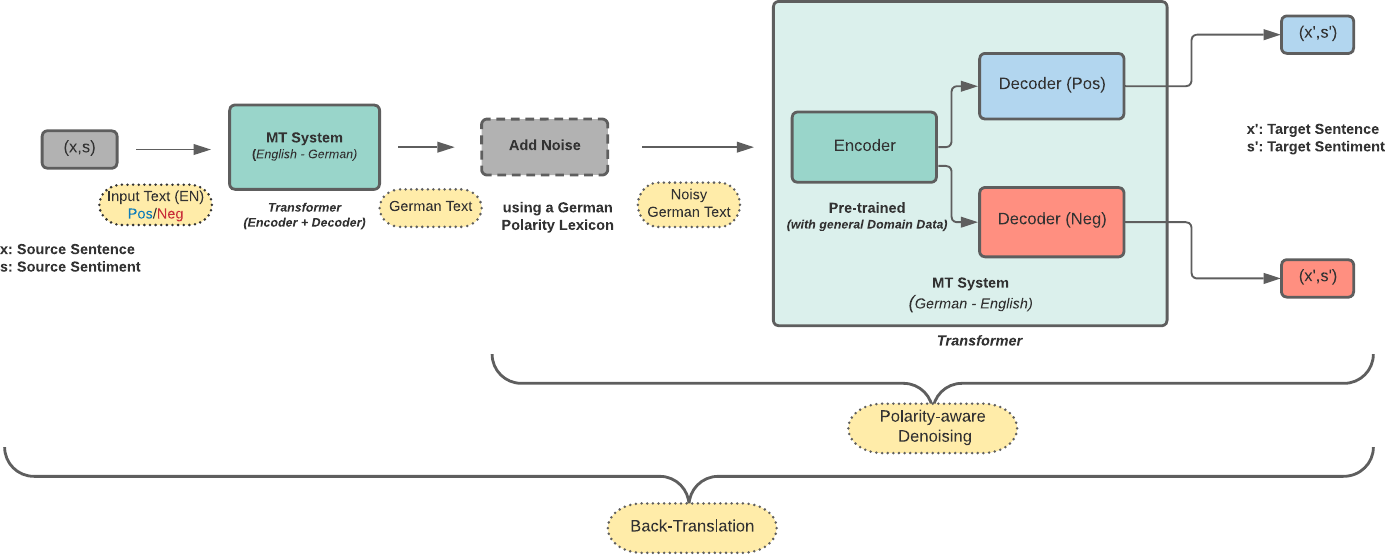}
    \caption{Our sentiment transfer pipeline. In the pipeline, we (1) \textit{translate} the source sentence from English to German using a transformer-based machine translation (MT) system; (2) \textit{apply noise} on the German sentence using a German polarity lexicon; (3) \textit{encode} the German sentence to latent representation using an encoder of German-to-English translation model; (4) \textit{decode} the shared latent representation using the decoder for the opposite sentiment.}
    \label{fig:proposed_method}
\end{figure*}

Figure \ref{fig:proposed_method} shows the overview of our proposed architecture. 
Following \citet{prabhumoye2018style}, we first translate the input text $x$ in the base language to a chosen intermediate language $\bar{x}$ using a translation model (see Section~\ref{sec:translation}).\footnote{We work with English as base language and German as intermediate language, see Section~\ref{sec:datasets}.} Next, we prepare a noisy text $x_{\textit{noise}}$ from $\bar{x}$ using polarity-aware noising with word deletion or masking probabilities $\theta_{N}$ (see Section~\ref{sec:pnoising}):
\begin{equation}
x_{\textit{noise}} = \textit{Noise}(\bar{x};\theta_{N}).    
\end{equation} 
We provide $x_{\textit{noise}}$ to the encoder of the $\bar{x}\rightarrow \hat{x}$ back-translation model (where $\hat{x}$ is a text in the base language with changed sentiment polarity). The encoder first converts the text to the latent representation $z$ as follows:
\begin{equation}
z = \textit{Encoder}(x_{\textit{noise}} ; \theta_E),
\end{equation}
where $\theta_E$ represent the parameters of the encoder.

Two separate sentiment-specific decoders are trained to decode the original positive and negative inputs by passing in their latent representations $z$:
\begin{align}
x_{\textit{pos}} &= \textit{Decoder}_{\textit{pos}}(z ; \theta_{D_{\textit{pos}}}) \\
x_{\textit{neg}} &= \textit{Decoder}_{\textit{neg}}(z ; \theta_{D_{\textit{neg}}}).
\end{align}

At inference time, sentiment transfer is achieved by decoding the shared latent representation using the decoder trained for the opposite sentiment, as follows:
\begin{align}
\hat{x}_{\textit{neg}} &= \textit{Decoder}_{\textit{pos}}(z ; \theta_{D_{\textit{pos}}}) \\
\hat{x}_{\textit{pos}} &= \textit{Decoder}_{\textit{neg}}(z ; \theta_{D_{\textit{neg}}})
\end{align}
where $\hat{x}_{neg}$, $\hat{x}_{pos}$ are the sentences with transferred sentiment conditioned on $z$ and $\theta_{D_{pos}}$ and $\theta_{D_{neg}}$ represent the parameters of the positive and negative decoders, respectively.

\section{Model Variants}
\label{sec:model-variants}

In all our experiments, we train sentiment transfer models using back-translation (Section~\ref{sec:translation}) based on the transformer architecture \cite{vaswani2017attention}. First, we present baselines for sentiment transfer with simple style conditioning (Section~\ref{sec:obaselines}). Next, we propose an approach based on an extended transformer architecture where we use separate modules (either the whole transformer model, or the transformer decoder only) for the respective target sentiment (Section~\ref{sec:obaselines}). We further improve upon our approach using polarity-aware denoising (Section~\ref{sec:pnoising}), which we propose as a new scheme for pre-training the sentiment transfer models.

\subsection{Back-translation} 
\label{sec:translation}
Back-translation for style transfer was introduced in \citet{prabhumoye2018style}. Following their approach, we use translation into German and subsequent encoding in a back-translation model to get a latent text representation for our sentiment transfer task. Prior work has also shown that the process of translating a sentence from a source language to a target language retains the meaning of the sentence but does not preserve the stylistic features related to the author’s traits \cite{rabinovich2016personalized}. A pure back-translation approach (without any specific provisions for sentiment) is referred to as \textit{Back-Translation} in our experiments.

We also experimented with an auto-encoder, but we have found that the back-translation model gives better results for sentiment transfer. We hypothesise that it is due to the fact that back-translation prevents the system from sticking to a particular wording, resulting in a more abstract latent representation.

\subsection{Our Baseline Models} 
\label{sec:obaselines}

In addition to a pure back-translation model, we present several straightforward baselines:

\begin{itemize}[itemsep=0pt,topsep=1pt,leftmargin=12pt]
\item\emph{Style Tok} is a back-translation model with added sentiment identifiers (\emph{<pos>} or \emph{<neg>}) as  output starting tokens. At the time of sentiment transfer, we decode the output with a changed sentiment identifier (\emph{<pos>} $\rightarrow$ \emph{<neg>}, \emph{<neg>} $\rightarrow$ \emph{<pos>}).


\item\emph{Two Sep. transformers:} \ To get more control over sentiment-specific generation, we train two separate transformer models for positive and negative sentiment, using only sentences of the respective target sentiment. During inference, the model is fed with inputs of the opposite sentiment, which it did not see during training. 

\item\emph{Shrd Enc + Two Sep Decoders:} \ We extend the above approach by keeping decoders separate, but using a shared encoder. During training, all examples are passed through the shared encoder, but each decoder is trained to only generate samples of one sentiment. Sentiment transfer is achieved by using the decoder for the opposite sentiment.

\item\emph{Pre Training Enc:} \ Following \citet{gururangan2020don}, we introduce a variant where the shared encoder is pretrained for back-translation on general-domain data. The pre-trained encoder is then further fine-tuned during sentiment transfer training. 

\end{itemize}

\subsection{Polarity-Aware Denoising} 
\label{sec:pnoising}

We devise a task-specific pre-training \cite{gururangan2020don} scheme for improving the sentiment transfer abilities of the model.
The idea of our pre-training scheme---\textit{polarity-aware denoising}---is to first introduce noise, i.e.\ delete or mask a certain proportion of words in the intermediate German input to the back-translation step, then train the model to remove this noise, i.e.\ produce the original English sentence with no words deleted or masked.
To decide which words get deleted or masked, we use automatically obtained sentiment polarity labels (see Section~\ref{sec:training-setup} for implementation details). This effectively adds more supervision to the task on the word level.
We apply three different approaches: deleting or masking (1) \textit{general} words (i.e., all the words uniformly), (2) \textit{polarity} words (i.e., only high-polarity words according to a lexicon), or (3) both \textit{general} and \textit{polarity} words (each with a different probability).

We use polarity-aware denoising during encoder pretraining, following the shared encoder and separate decoders setup from Section~\ref{sec:obaselines}. The encoder is further fine-tuned during the sentiment transfer training.



\section{Experiments}
\label{sec:experiments}

We evaluated and compared our approaches described in Section~\ref{sec:model-variants} with several state-of-the-art systems \cite{shen2017style, prabhumoye2018style, li2018delete, luo2019towards, wang2019controllable, he2020probabilistic} on two datasets (see Section~\ref{sec:datasets}).
We first describe our training setup (Section~\ref{sec:training-setup}), then detail our automatic evaluation metrics (Section~\ref{sec:autom-eval}) and human evaluation (Section~\ref{sec:human-eval}), and finally discuss the results (Section~\ref{sec:results}).

\subsection{Datasets}
\label{sec:datasets}

For our back-translation process and model pretraining, we used the WMT14 English-German \emph{(en-de)} dataset (1M sentences) from \citet{neidert2014stanford}.

For finetuning and experimental evaluation, we built a new English dataset for sentiment transfer, based on the Amazon Review Dataset \cite{ni2019justifying}. We have selected Amazon Review because it is more diverse topic-wise (books, electronics, movies, fashion, etc.) than existing sentiment transfer datasets, Yelp \cite{li2018delete} and IMDb \cite{dai2019style}, which are majorly focused on movie and restaurant/business-related reviews. 
For comparison with previous work, we also evaluate our models on the benchmark IMDb dataset \cite{dai2019style}.

While the Amazon Review data is originally intended for recommendation, it lends itself easily to our task. We have split the reviews into sentences and then used a pre-trained transformer-based sentiment classifier \cite{wolf-etal-2020-transformers} to select sentences with high polarity. Our intuition is that high-polarity sentences are more informative for the sentiment transfer task than neutral sentences. 
We filter out short sentences (less than 5 words) since it is hard to evaluate content preservation for these sentences.  We also ignored sentences with repetitive words (e.g., \emph{"no no no no thanks thanks."}) because these sentences are noisy and do not serve as good examples for the sentiment transfer model. 
We aim at comparable size to existing datasets \cite{li2018delete}: the resulting data has 102k positive and 102k negative examples in total, with 1+1k reserved for validation and testing  for each sentiment. The average sentence length in our data is 13.04 words.


\subsection{Training Setup}
\label{sec:training-setup}

In all our experiments, we use a 4-layer transformer \cite{vaswani2017attention} with 8 attention heads per layer. Both embedding and hidden layer size are set to 512. The same model shape is used for both the initial translation into German and the subsequent model handling back-translation, denoising, and sentiment transfer.

We use a German polarity lexicon to automatically identify pivot words for polarity-aware denoising. We prepared the German polarity lexicon by first translating the words from German to English using an off-the-shelf translation system \cite{lindattranslator}, followed by labeling the words with \textit{positive} and \textit{negative} labels using the English NLTK Vader lexicon \cite{hutto2014vader}. We performed a manual check of the results on a small sample. 

We test various combinations of noise settings w.r.t. noise probability, noise type (general or polarity-aware denoising), and noise mode (deleting or masking). Parameter values are pre-selected based on our preliminary experiments with the translation model (see Section~\ref{sec:translation}). The parameters are encoded in the name of the model as used in Table~\ref{tab:automatic_eval} (see the table caption for details).

\subsection{Automatic Evaluation Metrics}
\label{sec:autom-eval}

To evaluate the performance of the models, we compare the generated samples along three different dimensions using automatic metrics, following previous work: (1) style control, (2) content preservation, and (3) fluency. 

\begin{table*}[p]
\centering\small

\caption{\textbf{Automatic evaluation}. \textit{Accuracy}: Sentiment transfer accuracy. \textit{Sim} and \textit{B}: Cosine similarity and BLEU score between input and sentiment-transferred sentence. \textit{M/Sim} and \textit{M/B}: MaskSim and MaskBLEU (similarity and BLEU with polarity words masked, see Section~\ref{sec:autom-eval}). \textit{LM}: Average log probability assigned by vanilla GPT-2 language model. \textit{Avg}: Average length of transferred sentences. \textit{Avg}: Average of sentiment transfer accuracy, 100*MaskSim and MaskBLEU. 
Scores are based on a single run, with identical random seeds. 
First two sections show our own baselines, third section shows our models with denoising (with the best settings denoted SCT\textsubscript{1} and SCT\textsubscript{2}, see Section~\ref{sec:results}). The bottom section shows a comparison with state-of-the-art models.
Names of models with denoising reflect settings as follows: \textit{W} denotes WMT pretraining data, \textit{A} denotes Amazon finetuning data; the following tokens denote noise probability values are associated with the respective data. \textit{G/P} represents general/polarity token noising, \textit{D/M} represents noising mode deletion/masking. E.g. \textit{WG03P08-AG03P08-D}: noise probabilities on WMT and Amazon data are identical, noising by deletion on both general and polarity token noising is applied (with probabilities 0.3 and 0.8, respectively).}
\label{tab:automatic_eval}

\setlength{\tabcolsep}{4pt}
\renewcommand{\arraystretch}{0.8}
\begin{tabular}{lcccccccc}

\toprule
\textbf{Models} &  \textbf{Acc} & \textbf{Sim} & \textbf{M/Sim} & \textbf{B} & \textbf{M/B} & \textbf{LM} & \textbf{Len} & \textbf{Avg} \\

\toprule
\multicolumn{9}{c}{\textbf{Back-Translation Only (Section~\ref{sec:translation})}} \\
\midrule

\emph{Back-translation only} & \pz0.4 & 0.828 & 0.768 & 28.0 & 45.3 & -78.6 & 11.9 & 40.9 \\

\midrule
\multicolumn{9}{c}{\textbf{Our Baseline Models (Section~\ref{sec:obaselines})}} \\
\midrule

\textit{Style Tok} & 13.2 & 0.536 & 0.560 & \pz4.8 & \pz8.6 & -52.1 & \pz7.6 & 25.9 \\
\textit{Two Sep. transformers} & 89.3 & 0.394 & 0.611 & \pz6.8 & 19.6 & -79.0 & 13.7 & 56.7 \\
\textit{Shrd Enc + Two Sep Decoders} & 88.1 & 0.397 & 0.600 & \pz7.3 & 20.1 & -78.0 & 12.5 & 56.0 \\
\textit{Pre Training Enc} & 55.3 & 0.592 & 0.732 & 22.6 & 33.9 & -93.3 & 13.4 & 54.1 \\

\midrule
\multicolumn{9}{c}{\textbf{Our Models (with Denoising) (Section~\ref{sec:pnoising}) }} \\
\midrule

\textit{WG01-AG01-D} & 71.4 & 0.517 & 0.694 & 17.1 & 29.8 & -88.7 & 13.7 & 56.9 \\
\textit{WG01-AG01-M} & 68.0 & 0.536 & 0.711 & 19.4 & 31.1 & -86.3 & 12.6 & 56.7 \\
\textit{WG03-AG03-D} & 83.0 & 0.447 & 0.648 & 11.7 & 24.4 & -83.0 & 13.7 & 57.4 \\
\textit{WG03-AG03-M} & 78.8 & 0.481 & 0.669 & 14.2 & 28.2 & -82.7 & 13.0 & 58.0 \\
\midrule
\textit{WP08-AP08-D} & 66.9 & 0.528 & 0.701 & 19.5 & 31.3 & -82.8 & 12.4 & 56.1 \\
\textit{WP08-AP08-M} & 64.0 & 0.547 & 0.726 & 21.4 & 34.0 & -89.1 & 12.9 & 56.9 \\
\textit{WP1-AP1-D} & 58.7 & 0.570 & 0.727 & 22.7 & 33.1 & -87.2 & 12.2 & 54.8 \\
\textit{WP1-AP1-M} & 58.9 & 0.567 & 0.716 & 22.2 & 33.0 & -86.5 & 12.2 & 54.5 \\
\midrule
\textit{WG03-AG01-D} & 68.0 & 0.529 & 0.697 & 17.9 & 30.9 & -89.5 & 13.3 & 56.2 \\
\textit{WG03-AG01-M} & 80.7 & 0.473 & 0.665 & 13.9 & 27.5 & -82.8 & 13.1 & 58.2 \\
\textit{WG01-AG03-D} (=SCT\textsubscript{2}) & 85.2 & 0.441 & 0.646 & 11.8 & 25.4 & -79.8 & 13.1 & 58.4 \\
\textit{WG01-AG03-M} & 70.0 & 0.534 & 0.711 & 19.7 & 32.3 & -84.3 & 12.4 & 57.8 \\
\midrule
\textit{WP08-AP1-D} & 61.6 & 0.578 & 0.736 & 22.5 & 35.0 & -94.4 & 13.4 & 56.7 \\
\textit{WP08-AP1-M} & 60.9 & 0.554 & 0.724 & 22.0 & 33.3 & -85.5 & 12.6 & 55.6 \\
\textit{WP1-AP08-D} & 68.5 & 0.525 & 0.699 & 19.3 & 31.1 & -84.0 & 12.4 & 56.5 \\
\textit{WP1-AP08-M} & 61.1 & 0.560 & 0.714 & 21.5 & 32.9 & -86.0 & 12.1 & 55.1 \\
\midrule
\textit{WG03-AP08-D} & 67.0 & 0.533 & 0.697 & 20.3 & 31.7 & -84.3 & 12.5 & 56.1 \\
\textit{WG03-AP08-M} & 65.7 & 0.546 & 0.725 & 21.2 & 33.5 & -85.0 & 12.5 & 57.2 \\
\midrule
\textit{WP08-AG03-D} & 83.3 & 0.436 & 0.635 & 11.0 & 24.3 & -80.5 & 13.3 & 57.0 \\
\textit{WP08-AG03-M} & 79.6 & 0.473 & 0.665 & 13.2 & 26.9 & -83.1 & 13.2 & 57.6 \\
\midrule
\textit{WG03P08-AG03P08-D} & 65.5 & 0.547 & 0.705 & 20.3 & 32.6 & -90.4 & 13.2 & 56.2 \\
\textit{WG03P08-AG03P08-M} (=SCT\textsubscript{1}) & 82.0 & 0.460 & 0.665 & 13.7 & 27.4 & -79.6 & 12.8 & \textbf{58.6} \\

\midrule
\multicolumn{9}{c}{\textbf{State-of-the-Art Models}} \\
\midrule

\citet{shen2017style} & 88.6 & 0.346 & 0.513 & \pz3.2 & 18.3 & -74.0 & 10.9 & 52.7 \\
\citet{li2018delete} & 69.9 & 0.457 & 0.632 & 14.7 & 25.3 & -85.1 & 12.2 & 52.8 \\
\citet{luo2019towards} & 92.4 & 0.279 & 0.468 & \pz0.0 & \pz9.1 & -42.0 & \pz7.8 & 49.4 \\
\citet{prabhumoye2018style} & 93.5 & 0.308 & 0.504 & \pz0.9 & 15.2 & -61.0 & 10.3 & 53.0 \\
\citet{wang2019controllable} & 79.3 & 0.385 & 0.545 & 10.6 & 20.3 & -116.8 & 15.1 & 51.4 \\
\citet{he2020probabilistic} & 91.5 & 0.352 & 0.542 &\pz 9.5 & 21.8 & -65.9 & \pz8.2 & 55.8\\
\bottomrule
\end{tabular}
\end{table*}

\begin{table*}[t!]
\centering\small

\caption{Automatic evaluation on the IMDb Dataset (see Table~\ref{tab:automatic_eval} for metrics explanation).}
\label{tab:automatic_eval_imdb}

\setlength{\tabcolsep}{4pt}
\renewcommand{\arraystretch}{0.8}
\begin{tabular}{lcccccccc}

\toprule
\textbf{Models} &  \textbf{Acc} & \textbf{Sim} & \textbf{M/Sim} & \textbf{B} & \textbf{M/B} & \textbf{LM} & \textbf{Len} & \textbf{Avg} \\
\toprule
\citet{prabhumoye2018style}                & 87.1 & 0.345 & 0.480 & \pz2.7 & 14.3 & -63.5 & 10.0 &  49.8 \\
\citet{li2018delete}                       & 21.0 & 0.587 & 0.668 & 18.3 & 25.9 & -83.6 & 15.3 & 37.9 \\
\citet{wang2019controllable}               & 84.0 & 0.357 & 0.456 & \pz9.2 & 13.2 & -63.9 & 10.8 & 47.6 \\
\citet{he2020probabilistic}                & 81.7 & 0.458 & 0.576 & 29.0 & 41.8 & -83.6 & 15.3 & 60.4 \\
\midrule
SCT\textsubscript{1} (WG03P08-AG03P08-M)  & 85.3 & 0.435 & 0.612 & 28.6 & 42.3 & -86.4 & 15.9 & \textbf{62.9} \\
SCT\textsubscript{2} (WG01-AG03-D)        & 88.2 & 0.379 & 0.588 & 25.8 & 39.2 & -79.6 & 15.1 & 62.1 \\
\bottomrule
\end{tabular}
\end{table*}

\subsubsection{Standard Metrics}

\begin{itemize}[itemsep=0pt,topsep=1pt,leftmargin=12pt]
\item\emph{Style Accuracy:} \
Following prior work, we measure sentiment accuracy automatically by evaluating the sentiment of transferred sentences. We use a pre-trained transformer-based sentiment analysis pipeline\footnote{\url{https://huggingface.co/distilbert-base-uncased-finetuned-sst-2-english}} from Huggingface \cite{wolf-etal-2020-transformers}. 
    
\item\emph{Fluency:} \
We use the negative log-likelihood from the GPT-2 \cite{radford2019language} language model as an indirect metric for evaluating fluency. For context, we also calculate average sentence lengths of the sentiment-transferred sentences. 

\item\emph{Content Preservation:} \
Following previous work, we compute BLEU score \cite{papineni2002bleu} between the transferred sentence and its source. Higher BLEU indicates higher n-gram overlap between the sentences, which is generally viewed as proxy for content preservation. We also compute Sentence BERT \cite{reimers2019sentence} based cosine similarity score to match the vector space semantic similarity between the source and the transferred sentence. None of the techniques is capable of evaluating style transfer methods specifically with respect to preservation of content in style transfer \cite{toshevska2021review}. These metrics do not take into account the necessity of changing individual words while altering the sentence style. Intended differences between the source sentence and the transferred sentence are thus penalized. 

\end{itemize}

\subsubsection{Newly Introduced Metrics for Content Preservation}

To avoid the problems of the commonly used metrics, it makes sense in sentiment transfer to evaluate the content and similarity while ignoring any polarity tokens. Thus, we introduce MaskBLEU and MaskSim scoring methods -- these are identical to BLEU and cosine similarity, but they are computed on sentences where pivot words (based on NLTK Vader sentiment dictionary \cite{hutto2014vader}) have been masked. This allows measuring content preservation while ignoring the parts of the sentences that need to be changed.

\subsection{Human Evaluation}
\label{sec:human-eval}

As automated metrics for language generation do not correlate well with human judgements  \cite{novikova2017we}, we conduct an in-house human evaluation with five expert annotators. We randomly select 100 sentences (50 for each sentiment) from the our Amazon Review test set. The annotators rate model outputs on using a 1-5 Likert scale for style control, content preservation and fluency.

\subsection{Results}
\label{sec:results}

\subsubsection{Automatic Metrics} \label{subsec:autevalres}
\begin{figure}[t]
    \centering
    \includegraphics[width=0.50\textwidth]{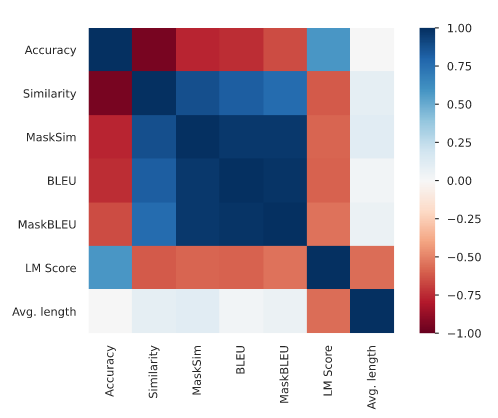}
    \caption{Correlations between automatic evaluation metrics on our Amazon Review data: sentiment accuracy is negatively correlated with BLEU, semantic similarity, and their masked variants.}
    \label{fig:corr}
\end{figure}

results on our Amazon Review data are shown in Table~\ref{tab:automatic_eval}. 
Overall, there is clearly a tradeoff between preserving sentiment-independent content and achieving the desired target sentiment.
Models which perform very well in sentiment transfer usually achieve worse results on content preservation.
This tradeoff is documented by correlations between the automatic metrics (see Figure~\ref{fig:corr}). Sentiment accuracy is negatively correlated with BLEU score, similarity measures as well as our newly introduced MaskBLEU and MaskSim scores.

\begin{table*}[t]
    \centering\small
    \caption{Human evaluation of sentiment transfer quality, content preservation, and fluency. Average of 1-5 Likert scale ratings on 100 examples from our Amazon Review data.}
    \label{tab:human_eval}
    \setlength{\tabcolsep}{2.1pt}
    \renewcommand{\arraystretch}{0.95}
    \begin{tabular}{ lccc }
    \midrule
     \textbf{Models} & \textbf{Sentiment} & \textbf{Content} & \textbf{Fluency} \\ 
     \midrule
     \citet{prabhumoye2018style} & 3.95 & 1.19 & 3.56 \\
     \citet{li2018delete} & 3.35 & 2.3 & 3.34 \\
     \citet{wang2019controllable} & 3.48 & 1.67 & 2.54 \\
     \citet{he2020probabilistic} & 3.69 & 1.66 & 3.26 \\
     \midrule
     SCT\textsubscript{1} (WG03P08-AG03P08-M) & 3.94 & \textbf{2.61} & 3.73 \\
     SCT\textsubscript{2} (WG01-AG03-D) & \textbf{3.99} & 2.56 & \textbf{3.79} \\
     \bottomrule
    \end{tabular}
\end{table*}


The translation-only and style token baselines do not perform well on changing the sentiment. Using two separate decoders leads to major sentiment transfer improvements, but content preservation is poor. Using the pre-trained encoder has helped to improve the content preservation, but sentiment transfer accuracy degrades significantly.

The main motivation for our work was to find a denoising strategy which offers the best balance between sentiment transfer and content preservation. Our results suggest putting an emphasis on denoising high-polarity words results in the best ratio between the sentiment transfer accuracy and content preservation metrics. Additionally, our models show the ability to produce fluent sentences, as shown by the language model score: our models' scores are similar to the back-translation baseline; other models only reach higher language model scores when producing very short outputs. 

Overall, our denoising approaches are able to balance well between sentiment transfer and content preservation. On content preservation, they perform much better than state-of-the-art models, and they stay competitive on style accuracy.
We selected two of our model variants -- SCT\textsubscript{1}=\textit{WG03P08-AG03P08-M} and SCT\textsubscript{2}=\textit{WG01-AG03-D} -- as the ones giving the best style-content trade-off (SCT) according to the average of sentiment accuracy, masked similarity and MaskBLEU (see Table~\ref{tab:automatic_eval}).



Automatic metrics on the IMDb dataset \cite{dai2019style} are shown in Table~\ref{tab:automatic_eval_imdb}, comparing our selected SCT\textsubscript{1} and SCT\textsubscript{2} models with state-of-the-art. Our models outperform the state-of-the-art in terms of sentiment accuracy and reach competitive results in terms of similarity, BLEU, and fluency. Same as on our Amazon Review data, they provide the best style-content trade-off (according to the averaged metric defined in Table~\ref{tab:automatic_eval}).

\subsubsection{Human Evaluation Results:} \label{subsec:humevalres}

We compare our best SCT\textsubscript{1} and SCT\textsubscript{2} models (selected above) with four state-of-the-art models: two of the most recent models \cite{wang2019controllable, he2020probabilistic}, and the models with best accuracy \cite{prabhumoye2018style} and MaskBLEU score \cite{li2018delete}.

We have evaluated over 600 model outputs. Results are presented in Table \ref{tab:human_eval}. The human evaluation results mostly agree with our automatic evaluation results. The results also show that our models are better in content preservation than the competitor models. 

We further examined a sample of the outputs in more detail to understand the behavior of different models.
We found that state-of-the-art models tend to lose the content of the source sentence, as shown in the example outputs in Table~\ref{tab:sample_output}. On the other hand, our models mostly preserve sentiment-independent content well while successfully transferring the sentiment. We conclude that with our models, there is a good balance between preserving the original sentiment-independent content and dropping the source sentiment, and existing state-of-the-art models tend to sacrifice one or the other.

\begin{table*}[t]
    \centering\small
    \caption{Example outputs comparison on samples from our Amazon Reviews dataset. Sentiment marker words (pivots) are colored. Note that our models preserve content better than most others.}
    \label{tab:sample_output}
    \setlength{\tabcolsep}{2.5pt}
    \renewcommand{\arraystretch}{0.8}
    \begin{tabular}{ p{3cm}|p{4.3cm}|p{4.3cm} } 
    \toprule
     \textbf{} & \textbf{\textcolor{red}{Negative} $\rightarrow$ \textcolor{blue}{Positive}} & \textbf{\textcolor{blue}{Positive} $\rightarrow$ \textcolor{red}{Negative}} \\ 
     \midrule
     \textbf{Source} & \hangindent=0.5em\textbf{movie was a \textcolor{red}{waste} of money : this movie totally \textcolor{red}{sucks} .} & \textbf{my daughter \textcolor{blue}{loves} them : )} \\
     \midrule
     \citet{prabhumoye2018style} & \hangindent=0.5em stan is always a \textcolor{blue}{great} place to get the food . & \textcolor{red}{do n't} be going here . \\
     \citet{li2018delete} & \hangindent=0.5em our \textcolor{blue}{favorite} thing was a movie story : the dream class roll ! & \hangindent=0.5em my daughter said i was still \textcolor{red}{not} acknowledged . \\
     \citet{wang2019controllable} & \hangindent=0.5em movie is a \textcolor{blue}{delicious} atmosphere of : this movie totally \textcolor{red}{sucks} movie ! & \hangindent=0.5em i should \textcolor{red}{not} send dress after me more than she would said \textcolor{red}{not} ? \\
     \citet{he2020probabilistic} & \hangindent=0.5em this theater was a \textcolor{blue}{great} place , we movie totally \textcolor{blue}{amazing} . & yup daughter has left ourselves . \\
     \midrule
     \raggedright\hangindent=3em SCT\textsubscript{1} (WG03P08-AG03P08-M) & \hangindent=0.5em movie : a \textcolor{blue}{great} deal of money : this movie is absolutely \textcolor{blue}{perfect} . &  \hangindent=0.5em my daughter \textcolor{red}{hates} it : my daughter .\\
     SCT\textsubscript{2} (WG01-AG03-D) & this movie is a \textcolor{blue}{great} deal of money. & my daughter \textcolor{red}{hated} it . \\
          
     \midrule
     \textbf{Source} & \hangindent=0.5em\textbf{\textcolor{red}{nothing} truly interesting happens in this book .} &  \hangindent=0.5em\textbf{\textcolor{blue}{best} fit for my baby : this product is \textcolor{blue}{wonderful} ! !} \\
     \midrule
     \citet{prabhumoye2018style} & very \textcolor{blue}{good} for the \textcolor{blue}{best} . &  \hangindent=0.5em \textcolor{red}{bad} customer service to say the food , and it is n't . \\
     \citet{li2018delete} & \hangindent=0.5em \textcolor{red}{nothing} truly interesting happens in this book . &  \hangindent=0.5em my mom was \textcolor{red}{annoyed} with my health service is no notice . \\
     \citet{wang2019controllable} & \hangindent=0.5em \textcolor{red}{nothing} truly interesting happens in this book make it casual and spot . &  \hangindent=0.5em do not buy my phone : this bad crap was worst than it ? \\
     \citet{he2020probabilistic} & \hangindent=0.5em haha truly interesting happens in this book . &  \textcolor{red}{uninspired} . \\
     \midrule
     \raggedright\hangindent=3em SCT\textsubscript{1} (WG03P08-AG03P08-M) & \hangindent=0.5em in this book is truly a really \textcolor{blue}{great} book . & \hangindent=0.5em \textcolor{red}{not good} for my baby : this product is \textcolor{blue}{great} ! ! ! ! ! ! ! ! \\
     SCT\textsubscript{2} (WG01-AG03-D) & in this book is truly \textcolor{blue}{awesome} . &  \hangindent=0.5em \textcolor{red}{not happy} for my baby : this product is \textcolor{red}{not great} ! ! \\
     

     \bottomrule
    \end{tabular}
\end{table*}

\section{Conclusions and Future Work}

We proposed an approach for text sentiment transfer based on the transformer architecture and polarity-aware denoising. Experimental results on two datasets showed that our method achieves competitive or better performance compared to state-of-the-art. 
Our architecture provides a good style-content tradeoff mainly due to two elements: (1)~separate sentiment-specific decoders providing explicit target sentiment control, and (2)~polarity-aware enhanced denoising removing sentiment implicitly at the token level. 
As shown by human evaluation and our manual inspection, our models still sometimes fail to preserve the meaning of the original. While we improve upon previous works on content preservation, this remains a limitation.

In the future, we plan to adapt our method to the different kind of style transfer tasks such as formality transfer or persona-based text generation. Lexicons for the required attribute makers can be extracted by mining stylistic markers from generic dictionaries, or from personality-annotated data \cite{mairesse_controlling_2011}.
We also intend to focus on better controlling content preservation with the use of semantic parsing.

\section*{Acknowledgments}

This research was supported by Charles University projects GAUK 392221, GAUK 140320, SVV 260575 and PRIMUS/19/SCI/10, and by the European Research Council (Grant agreement No. 101039303 NG-NLG). It used resources provided by the LINDAT/CLARIAH-CZ Research Infrastructure (Czech Ministry of Education, Youth and Sports project No. LM2018101).

%
%
%
\bibliographystyle{splncs04nat}
\bibliography{references}
%




\end{document}